%% file: anonymous-submission-latex-2023.tex
\documentclass[letterpaper]{article} 
\usepackage{aaai23}  
\usepackage{times}  
\usepackage{helvet}  
\usepackage{courier}  
\usepackage[hyphens]{url}  
\usepackage{graphicx} 
\usepackage{natbib}  
\usepackage{caption} 
\usepackage{algorithm}
\usepackage{algorithmic}
\usepackage[table]{xcolor}
\usepackage{tikz}
\usepackage{comment}
\usepackage{amsmath,amssymb} 
\usepackage{color}
\usepackage{url}            
\usepackage{booktabs}       
\usepackage{amsfonts}       
\usepackage{comment}
\usepackage{color}
\usepackage{multirow}
\usepackage{array}
\usepackage{graphbox}
\usepackage{booktabs} 
\usepackage{wrapfig}
\usepackage{enumitem}
\usepackage{bbding}

\usepackage{newfloat}
\usepackage{listings}
\DeclareCaptionStyle{ruled}{labelfont=normalfont,labelsep=colon,strut=off} 
\floatstyle{ruled}
\newfloat{listing}{tb}{lst}{}
\floatname{listing}{Listing}
\newcommand{\ie}[1]{\textit{i.e.}{#1}}
\newcommand{\eg}[1]{\textit{e.g.}{#1}}
\newcommand{\mypara}[1]{\vspace{0.2cm}\noindent\textbf{#1}\hspace{0.1cm}}
\renewcommand{\vec}[1]{\boldsymbol{#1}}

\title{Learning Context-Aware Classifier for Semantic Segmentation}

\author{
	Zhuotao Tian$^{1,4}$,
	Jiequan Cui$^{1}$, 
	Li Jiang$^{2}$, 
	Xiaojuan Qi$^{3}$, 
	Xin Lai$^{1}$\\
	Yixin Chen$^{1}$, 
	Shu Liu$^{4}$, 
	Jiaya Jia$^{1,4}$
}
\affiliations{
$^{1}$The Chinese University of Hong Kong \\
$^{2}$Max Planck Institute for Informatics \\
$^{3}$The University of Hong Kong \\
$^{4}$SmartMore Corporation \\
}

\usepackage{bibentry}

\begin{document}

\maketitle

\input{doc/abstract}	

\input{doc/intro}

\input{doc/relatedwork}

\input{doc/method}

\input{doc/results}

\input{doc/conclusion}

\section*{Acknowledgements}
This work is supported by Shenzhen Science and Technology Program (KQTD20210811090149095).

{
	\bibliographystyle{ieee_fullname}
	\bibliography{egbib}
}

\end{document}

%% file: doc/abstract.tex
\begin{abstract}
Semantic segmentation is still a challenging task for parsing diverse contexts in different scenes, thus the fixed classifier might not be able to well address varying feature distributions during testing.  Different from the mainstream literature where the efficacy of strong backbones and effective decoder heads has been well studied, in this paper, additional contextual hints are instead exploited via learning a context-aware classifier whose content is data-conditioned, decently adapting to different latent distributions. Since only the classifier is dynamically altered, our method is model-agnostic and can be easily applied to generic segmentation models. Notably, with only negligible additional parameters and +2\% inference time, decent performance gain has been achieved on both small  and large models  with challenging benchmarks, manifesting substantial practical merits brought by our simple yet effective method. The implementation is available at \url{https://github.com/tianzhuotao/CAC}.

\end{abstract}

%% file: doc/intro.tex
\section{Introduction}
As a fundamental tool, semantic segmentation has profited a wide range of applications~\cite{zhang2022mediseg,tian2019lase}.
Recent advances regarding model structure for boosting segmentation performance are fastened to stronger backbones and decoder heads, focusing on delicate designs to yield high-quality features, and then they all apply a classifier to make predictions.

\begin{figure}[!t]
	\centering
	\begin{minipage}   {0.72\linewidth}
		\centering
		\includegraphics [width=1\linewidth] 
		{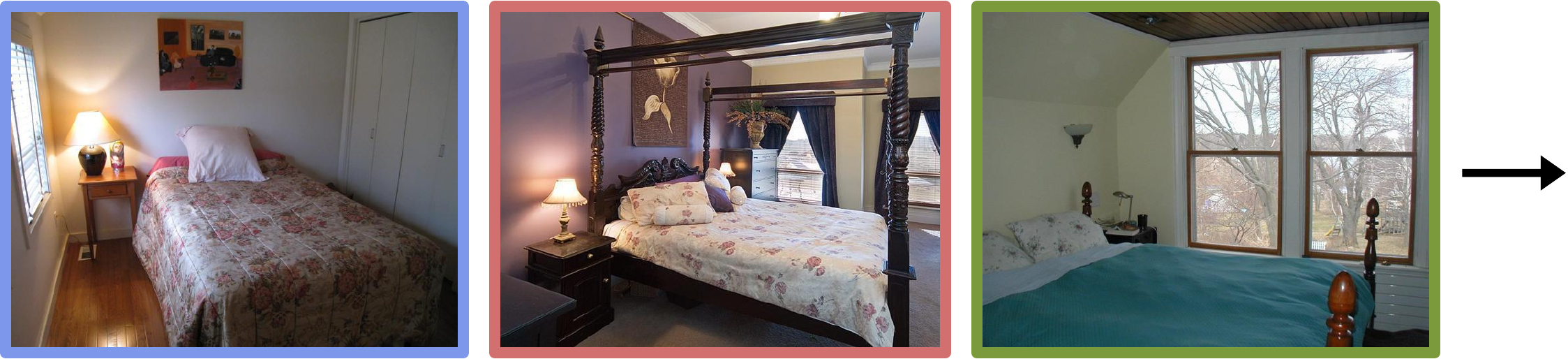}
	\end{minipage}       
	\begin{minipage}   {0.24\linewidth}
		\centering
		\includegraphics [width=1\linewidth ,height=0.7\linewidth]
		{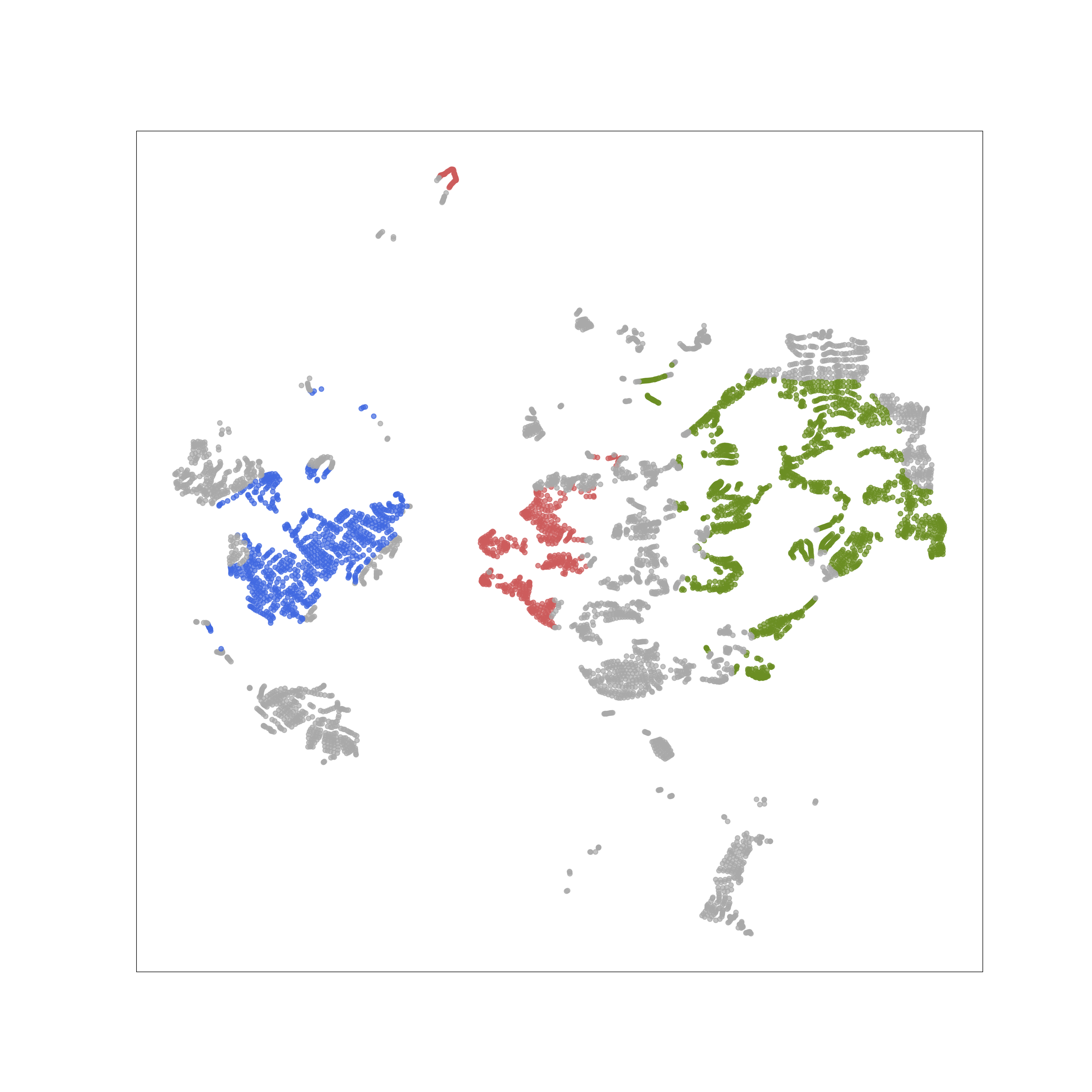}
	\end{minipage}              
	\caption{Visualizations of latent features of \textit{Bed} in different scenes. \textit{Red}, \textit{blue} and \textit{green} represent features belonging to \textit{Bed} in the left three images respectively, and gray denotes the embeddings of the other co-occurring classes.}
	\label{fig:init_tsne}
\end{figure}

However, the classifier in the recent literature is composed of a set of parameters shared by all images, leading to an inherent challenge during testing that the fixed parameters are required to handle diverse contexts contained in various samples with different co-occurring objects and scenes, e.g., domain adaptation~\cite{lai2021cac} Even for pixels in the same category, embeddings from different images cannot be well clustered as shown in Figure~\ref{fig:init_tsne}, potentially inhibiting the segmentation performance with the fixed classifier. This observation induces a pertinent question: \textit{whether the classifier can be enriched with contextual information for individual images.}

Consequently, in this paper, we attempt to yield context-aware classifier whose content is data-conditioned, decently describing different latent distributions and thence making accurate predictions.
To investigate the feasibility, we start from an ideal scenario where the precise contextual hints are provided to the classifier by the ground-truth label that enables forming perfect categorical feature prototypes to supplement the original classifier. 
As illustrated in Figure~\ref{fig:tsne_vis}, the classifier enriched with impeccable contextual priors significantly outperforms the baseline in both training and testing phases, certifying the superior performance upper bound achieved by the context-aware classifier. 

Yet, ground-truth label is not available during testing; therefore, in an effort to approximate the aforementioned oracle situation, we instead let the model learn to yield the context-aware classifier by mimicking the predictions made by the oracle counterpart. 
Nevertheless, treating elements equally during the imitation process is found deficient, in that the informative cues may be suppressed by those not instructive. To alleviate this issue, the class-wise entropy is leveraged to accommodate the learning process. 

\begin{figure}[!t]
	\centering
	\begin{minipage}   {0.48\linewidth}
		\centering
		\includegraphics [width=1\linewidth ,height=0.8\linewidth]
		{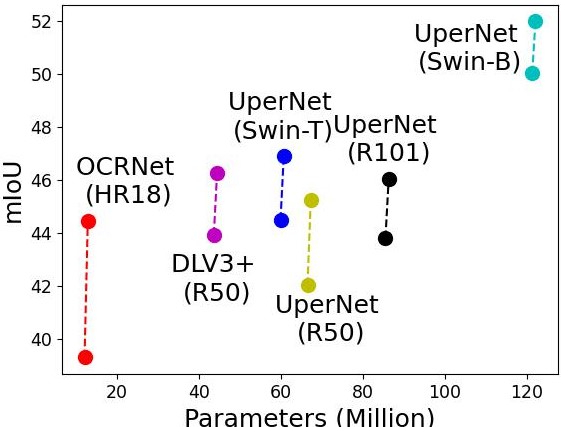}	
	\end{minipage}    
	\begin{minipage}   {0.48\linewidth}
		\centering
		\includegraphics [width=1\linewidth ,height=0.8\linewidth]
		{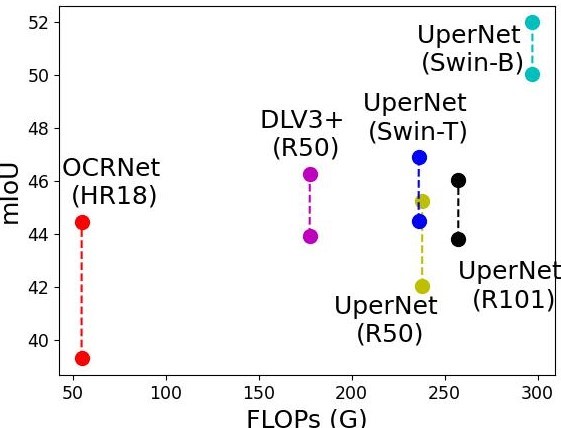}
	\end{minipage}        
	\caption{Effects on model performance (mIoU) and efficiency (parameters and inference time) on ADE20K~\cite{ade20k}. Detailed results are shown in Table~\ref{tab:results_ade20k}. }
	\label{fig:statistics}
\end{figure}

The proposed method is model-agnostic, thus it can be applied to a wide collection of semantic segmentation models with generic encoder-decoder structures. 
To this end, with our method, as shown in Figure~\ref{fig:statistics}, significant performance gains have been constantly brought to both small and large models without compromising the model efficiency, \ie, only about 2\% increase on inference time and a few additional parameters, even boosting the small model OCRNet (HR18)~\cite{ocnet,hrnet_cvpr} to reach higher performance than the competitors with much more parameters. 
To summarize, our contributions are as follows. 
\begin{itemize}
    \item We propose to learn the context-aware classifier whose content varies according to different samples, instead of a static one used in common practice.
    \item To make the context-aware classifier learning tractable, an entropy-aware KL loss is designed to mitigate the adverse effects brought by information imbalance. 
    \item Our method is easy to be plugged into other existing segmentation models, achieving considerable improvement with little compensation on efficiency. 
    
\end{itemize}

%% file: doc/relatedwork.tex
\section{Related Work}
Semantic segmentation is a fundamental yet challenging task where precise pixel-wise predictions are needed. However, models cannot make prediction for each position merely based on its RGB values, thus broader contextual information are exploited to achieve decent performance.

FCN~\cite{fcn} proposes to adopt the convolution layers to tackle the semantic segmentation task. Then, well-designed decoders~\cite{deconvnet,segnet,unet} are proposed to gradually up-sample the encoded features in low resolution, so as to retain sufficient spatial information for yielding accurate predictions. 
Besides, since the receptive field is important for scene parsing, dilated convolutions~\cite{deeplab,dilation}, global pooling~\cite{parsenet} and pyramid pooling~\cite{deeplab,pspnet,denseaspp,tian2020pfenet,strip} are proposed for further enlarging the receptive field and mining more contextual cues from the latent features extracted by the backbone network. More recently, pixel and region contrasts are exploited~\cite{pixelcontrast,lai2021cac,regioncontrast,jiang2021semi,cui2022generalized}. 

Also, transformer performs dense spatial reasoning, thus it is adopted in decoders for modelling the long-range relationship in the extracted features~\cite{ocnet,psanet,catrans,cui2022region,encnet,danet,ccnet,ocr,maskformer}.  Transformer-based backbones take a step further because the global context can be modeled in every layer of the transformer, achieving new state-of-the-art results. Concretely, by applying a pure transformer ViT~\cite{vit} as the feature encoder, \cite{setr,segmenter} set up new records on semantic segmentation against the other convolution-based competitors, and Swin Transformer~\cite{swin} further manifests the superior performance with the decoder head of UperNet~\cite{upernet}. Besides, SegFormer~\cite{segformer} is a framework specifically designed for segmentation by combining both local and global attentions to yield informative representations. 

In summary, the mainstream of research aiming at improving segmentation model structures focuses on either designing backbones for feature encoding or developing decoder heads for producing informative latent features, and the classifier is seldom studied. Instead, we exploit the semantic cues in individual samples via learning to form the context-aware classifiers, keeping the rest intact.

%% file: doc/method.tex
\section{Our Method}
\label{sec:method}

\subsection{Motivation}
A generic deep model can be deemed as a composition of two modules: 1) feature generator $\mathcal{G}$ and 2) classifier $\mathcal{C}$. The feature generator $\mathcal{G}$ receives the input image $x$ and projects it into high-dimensional feature $\vec{f} \in \mathcal{R}^{[hw \times d]}$ where $h$, $w$ and $d$ denote the height, width and dimension number of the feature $\vec{f}$, respectively. Necessary contextual information is enriched in the extracted feature by the feature generator, ensuring the classifier $\mathcal{C}\in \mathcal{R}^{[n \times d]}$ can make prediction $\vec{p}\in \mathcal{R}^{[hw \times n]}$ for $n$ classes on different positions individually. Put differently, the aforementioned process implies that the classifier should serve as a feature descriptor whose weights are used as decision boundaries in the high-dimensional feature space, decently describing the feature distribution and making the judgment, \ie, pixel-wise predictions.

\begin{figure}[!t]
	\centering
	\begin{minipage}   {0.95\linewidth}
		\centering
		\includegraphics [width=0.95\linewidth,height=0.65\linewidth]
		{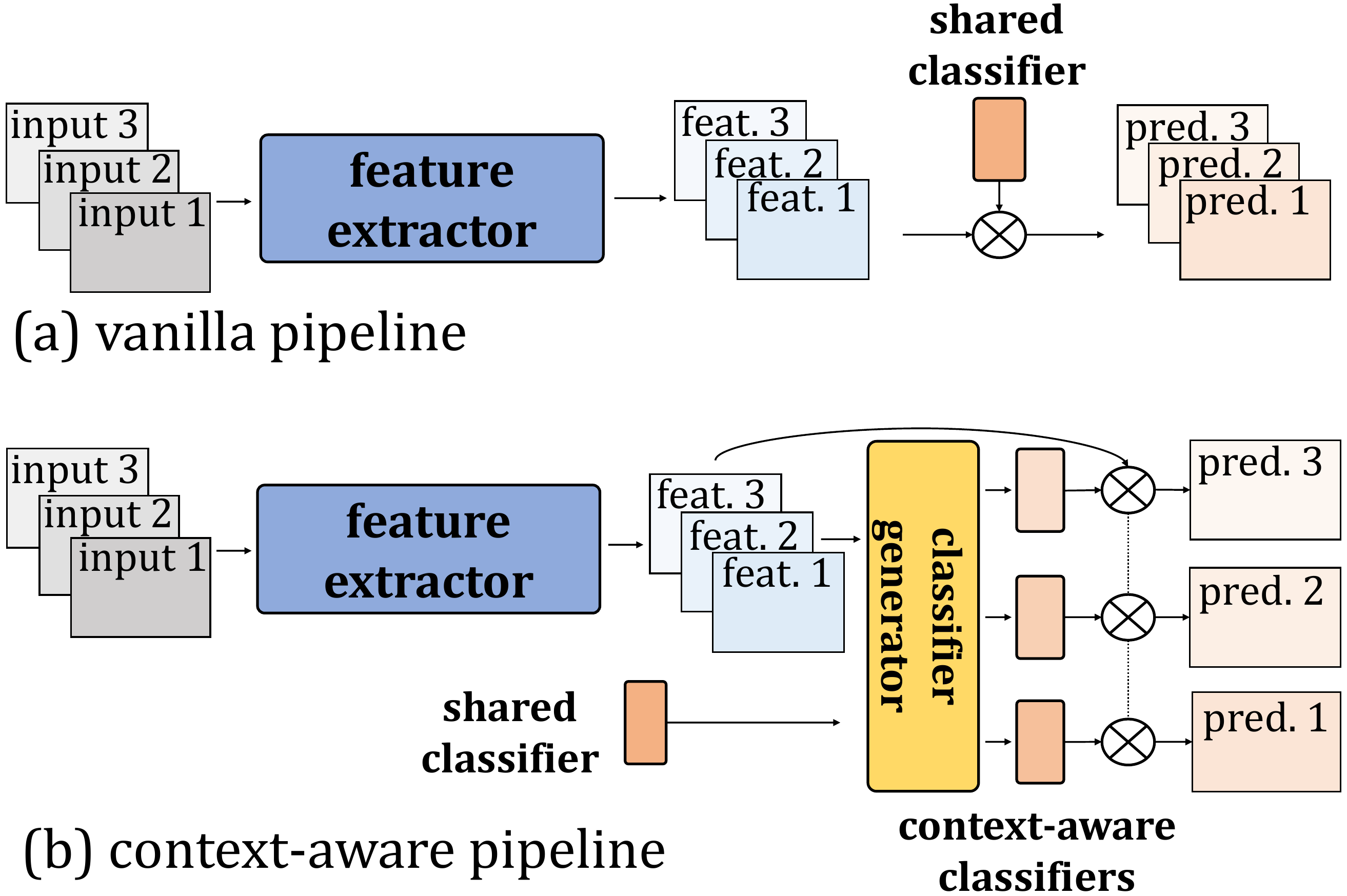}
	\end{minipage}     
	\caption{Comparison between (a) the vanilla and (b) our proposed pipelines. }
	\label{fig:overview}
\end{figure}

However, images for semantic segmentation usually have distinct contextual hints, thus we conjecture that using the universal feature descriptor, \ie, classifier, shared by all testing samples might not be the optimal choice for parsing local details for the individual ones. This inspires us to explore a feasible way by which the classifier becomes ``context-aware'' to different samples, improving the performance but keeping the structure of the feature generator intact, as abstracted in Figure~\ref{fig:overview}. 

\subsection{Is Context-Aware Classifier Necessary? }
\label{sec:oracle_case}
With an eye towards enriching contextual cues to the classifier, essential information should be mined from the extracted features. To verify the hypothesis that the proposed context-aware classifier is conducive to the model performance, we start with a case study regarding the oracle situation where the contextual information is exactly enriched with the guidance of ground-truth annotation that can offer precise contextual prior.

Specifically, given the extracted feature map $\vec{f} \in \mathcal{R}^{[hw \times d]}$ and the vanilla classifier $\mathcal{C}\in \mathcal{R}^{[n \times d]}$ of $n$ classes, the pixel-wise ground-truth annotation $\vec{y} \in \mathcal{R}^{[hw]}$ can be accordingly transformed into $n$ binary masks $\vec{y}_{*} \in \mathcal{R}^{[n \times hw]}$ 
indicating the existence of $n$ classes in $\vec{y}$. 
Then, we can obtain the categorical prototypes $\mathcal{C}_y \in \mathcal{R}^{[n, d]}$ by applying masked average pooling (MAP) with $\vec{y}_{*}$ and $\vec{f}$:
\begin{equation}
    \label{eq:oracle_proto}
    \mathcal{C}_y = \frac{\vec{y}_{*} \times \vec{f}}{\sum_{j=1}^{hw} \vec{y}_{*}(\cdot, j)}.
\end{equation}
Then, the oracle context-aware classifier $\mathcal{A}_{y} \in \mathcal{R}^{[n, d]}$ is yielded by taking the merits from both $\mathcal{C}_y$ and $\mathcal{C}$  with a light-weight projector $\theta_y$ that is composed of two linear layers. This process can be expressed as
\begin{equation}
    \label{eq:oracle_cls}
    \mathcal{A}_y = \theta_y(\mathcal{C}_y \oplus \mathcal{C}),
\end{equation}
where $\oplus$ denotes the concatenation on the second dimension. An alternative choice is simply adding $\mathcal{C}_y$ and $\mathcal{C}$, while experimental results in Table~\ref{tab:ablation_alternative_designs} show that concatenation with projection leads to better performance. Finally, the prediction $\vec{p}_{y}$ obtained with the oracle context-aware classifier $\mathcal{A}_y$ is yielded as:
\begin{equation}
    \label{eq:oracle_pred}
    \vec{p}_{y} = \tau \cdot \eta(\vec{f}) \times \eta(\mathcal{A}_y)^\top,
\end{equation}
where $\eta$ is the L-2 normalization operation along the second dimension, thus Eq.~\eqref{eq:oracle_pred} is calculating the cosine similarities. $\tau$ scales the output value range from [-1,1] to [-$\tau$, $\tau$], so that $\vec{p}_{y}$ can be decently optimized by the standard cross-entropy loss. We empirically set $\tau$ to 15 in experiments. The necessity of cosine similarity and sensitivity analysis regarding  $\tau$ are discussed in Section~\ref{sec:ablation_study}.

\begin{figure}[!t]
	\centering     
    \begin{minipage}   {0.49\linewidth}
        \centering
        \includegraphics [width=1\linewidth,height=0.8\linewidth]
        {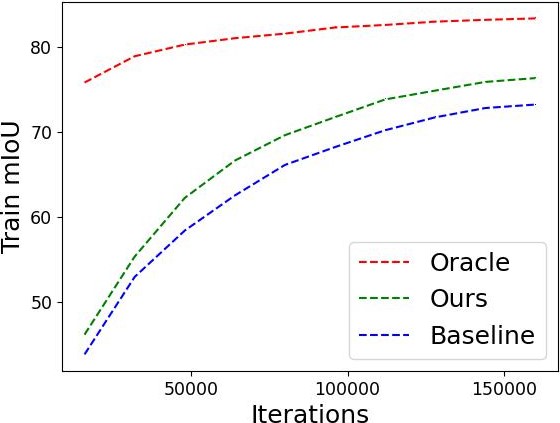}
    \end{minipage} 
    \begin{minipage}   {0.49\linewidth}
        \centering
        \includegraphics [width=1\linewidth,height=0.8\linewidth]
        {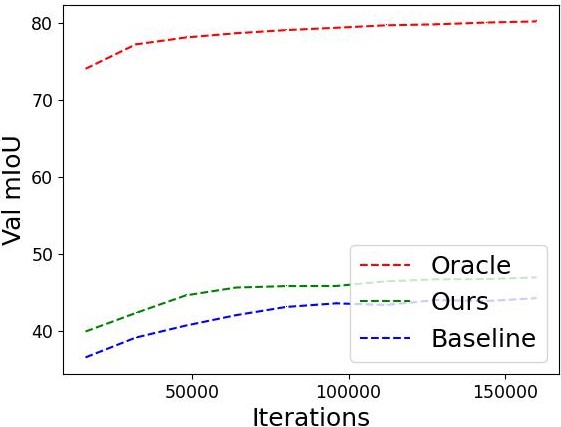}
    \end{minipage}    
	\footnotesize
	\caption{Visual comparison between \textit{train} (left) and \textit{val} (right) mIoU curves. Results are obtained with UperNet+Swin-Tiny~\cite{upernet,swin} on ADE20K~\cite{ade20k}. }
	\label{fig:tsne_vis}
\end{figure}

\mypara{Results and discussion. }
As shown by the red and blue curves in Figure~\ref{fig:tsne_vis}, by simply substituting the original classifier $\mathcal{C}$ with the oracle context-aware classifier $\mathcal{A}_{y}$, samples of different classes can be better distinguished via a better feature descriptor serving as the decision boundary. It implies that additional detailed co-occurring semantic cues conditioned on individual testing samples have been exploited by $\mathcal{A}_{y}$, so as to achieve preferable performance on both training and validation sets during both training and testing phases.

However, $\mathcal{A}_{y}$ is obtained with the ground-truth annotation that is only available during model training. To make it tractable for boosting the testing performance, learning to form such  context-aware classifiers conditioned on the content of individual samples takes the next step.

\begin{figure}[!t]
	\centering
	\begin{minipage}   {0.95\linewidth}
		\centering
		\includegraphics [width=1\linewidth] 
		{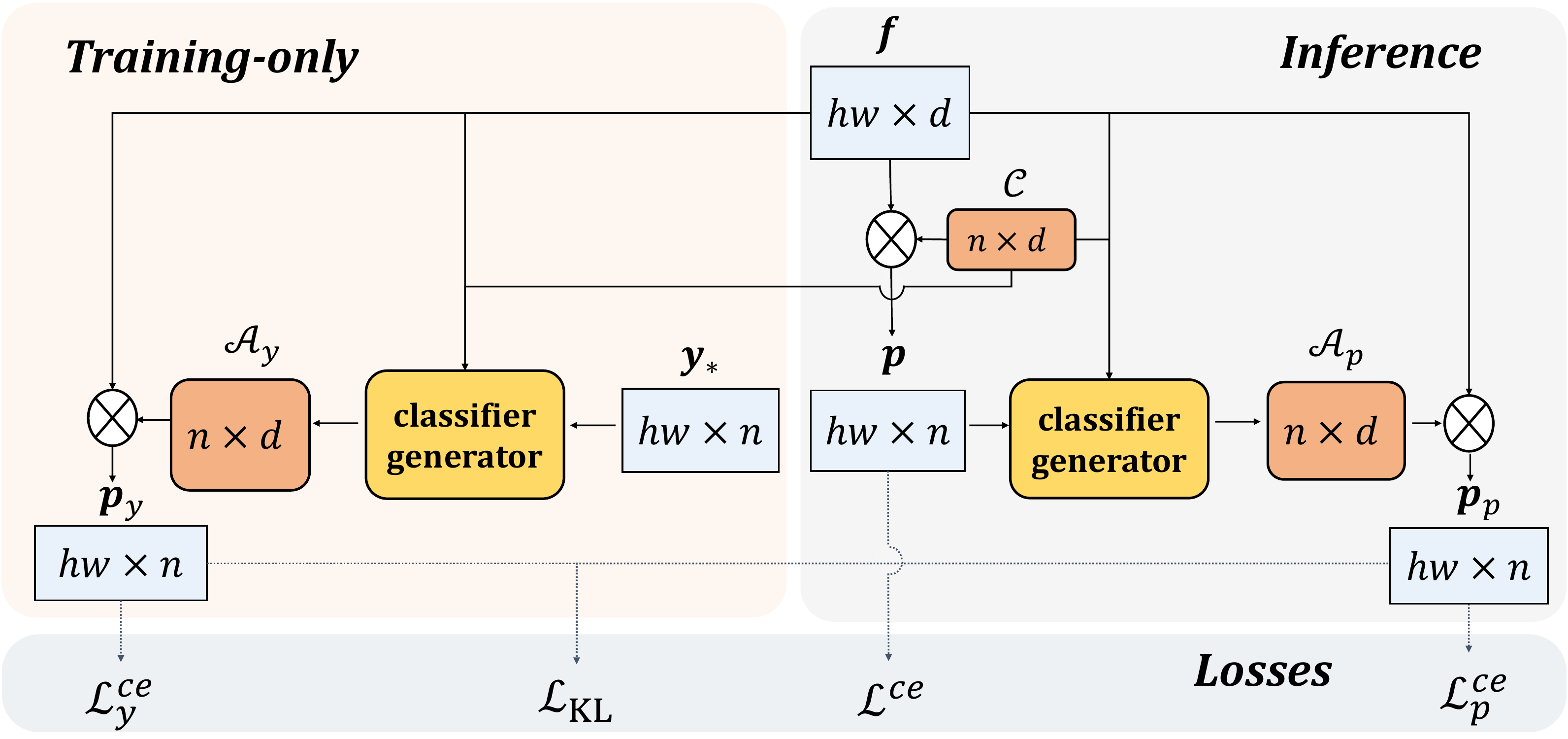}
	\end{minipage}     
	\caption{Pipeline for learning context-aware classifier. }
	\label{fig:detail}
\end{figure}

\subsection{Learning Context-Aware Classifier}
\label{sec:learning_cwc}
Without ground-truth labels, a natural modification to the oracle case is to use prediction $\vec{p}$ instead of ground-truth label $\vec{y}_{*}$ to approximate the oracle contextual prior. The overall learning process is illustrated in Figure~\ref{fig:detail}. 

Specifically, we note that the prediction $\vec{p}\in \mathcal{R}^{[hw \times n]}$ refers to the results got with the original classifier, \ie,  $\vec{p} = \vec{f} \times \mathcal{C}^\top$.  Therefore, the estimated contextual prototypes $\mathcal{C}_p \in \mathcal{R}^{[n \times d]}$ are yielded with $\vec{p}$ as  
\begin{equation}
    \label{eq:pred_proto}
    \mathcal{C}_p = \frac{\sigma(\vec{p})^\top \times \vec{f}}{\sum_{j=1}^{hw} \sigma(\vec{p})^\top(\cdot, j)} 
                  = \frac{\sigma(\vec{f} \times \mathcal{C}^\top)^\top \times \vec{f}}{\sum_{j=1}^{hw} \sigma(\vec{f} \times \mathcal{C}^\top)^\top(\cdot, j)},
\end{equation}
where $\sigma$ is Softmax operation applied on the second dimension.
Similar to Eq.~\eqref{eq:oracle_cls}, the context-aware classifier $\mathcal{A}_p  \in \mathcal{R}^{[n, d]}$ is yielded by processing the concatenation of the estimated contextual prior $\mathcal{C}_p$ and the original classifier $\mathcal{C}$ as shown in Eq.~\eqref{eq:estimate_cls}:
\begin{equation}
    \label{eq:estimate_cls}
    \mathcal{A}_p = \theta_p(\mathcal{C}_p \oplus \mathcal{C}),
\end{equation}
where $\theta_p$ denotes the projector that has the same structure as $\theta_y$. Also, prediction $\vec{p}_{p}$ represents the result got from the temporarily estimated context-aware classifier $\mathcal{A}_p$ as shown in Eq~\eqref{eq:estimate_pred}. 
\begin{equation}
    \label{eq:estimate_pred}
    \vec{p}_{p} = \tau \cdot \eta(\vec{f}) \times \eta(\mathcal{A}_p)^\top.
\end{equation}
We find that adopting the context-aware classifier to calculate the cosine similarities yields better results than the commonly used dot product, because the former helps alleviate the issues that stem from the instability of the individually generated $\mathcal{A}_y$ and $\mathcal{A}_p$. Contrarily, simply replacing the dot product used by the original classifier with cosine similarity is not profitable to the overall performance. More detailed discussions and experiments are shown in Section~\ref{sec:ablation_study}. 

\mypara{Optimization. }
Using a single pixel-wise cross-entropy (CE) loss $\mathcal{L}_{p}^{ce}$ to supervise $\vec{p}_{p}$ seems feasible for learning the context-aware classifier. However, as shown in later experiments in Table~\ref{tab:ablation_loss}, standard CE loss brings incremental improvement to the baseline because, compared to the precise prior offered by the ground-truth $\vec{y}$, the uncertainty contained in $\vec{p}$ makes the estimated categorical prototypes $\mathcal{C}_p$ less reliable than the universally shared  classifier $\mathcal{C}$, potentially making the projector $\theta_p$ tend to trivially neglect $\mathcal{C}_p$.

As discussed in Section~\ref{sec:oracle_case}, the oracle context-aware classifier $\mathcal{A}_y$ yielded with ground-truth label is a better distribution descriptor for each sample, thus it achieves much better performance than the original classifier $\mathcal{C}$. 
Therefore, inspired by the practices in knowledge distillation~\cite{kd_hinton} and incremental learning~\cite{lwf}, as a means to transfer or retain necessary information, we additionally incorporate KL divergence $\mathcal{L}_{KL}$ to regularize the model such that it is encouraged to yield more informative $\mathcal{A}_p$ by mimicking the prediction $\vec{p}_{y}$ of the oracle situation $\mathcal{A}_y$. In other words, useful knowledge is distilled from $\mathcal{A}_y$ to $\mathcal{A}_p$:
\begin{equation}
    \label{eq:init_kl_loss}
    \mathcal{L}_{KL} = -\frac{1}{hw} \sum_{i=1}^{hw} \sum_{j=1}^{n} \sigma(\vec{p}_{y})^{i,j} \cdot \log \sigma(\vec{p}_{p})^{i,j},
\end{equation}
where $h$, $w$ and $n$ denote height, width and class number, and $\sigma$ represents the Softmax operation applied to the second dimension of $\vec{p}_{y}\in \mathcal{R}^{[hw, n]}$ and $\vec{p}_{p}\in \mathcal{R}^{[hw, n]}$. Gradients yielded by $\mathcal{L}_{KL}$ will not be back-propagated to $\vec{p}_{y}$. 

In addition to $\mathcal{L}_{p}^{ce}$ and $\mathcal{L}_{KL}$, CE losses applied to $\vec{p}$ and $\vec{p}_y$, denoted as $\mathcal{L}^{ce}$ and $\mathcal{L}_{y}^{ce}$ respectively, are also optimized, intending to ensure the quality of the estimated and the oracle prototypes. 
To this end, the training objective $\mathcal{L}$ is:
\begin{equation}
    \label{eq:init_loss}
    \mathcal{L} = \mathcal{L}^{ce} + \mathcal{L}_{p}^{ce} + \mathcal{L}_{y}^{ce} + \lambda_{KL}\mathcal{L}_{KL}.
\end{equation}

\mypara{Entropy-aware distillation. }
The KL divergence $\mathcal{L}_{KL}$ introduced in Eq.~\eqref{eq:init_kl_loss} distills the categorical information from $\mathcal{A}_y$ to $\mathcal{A}_p$, so as to let the model learn to approximate the oracle case.  Also, for segmentation, the one-hot label is not always semantically accurate because it cannot reveal the actual categorical hints in each image, but soft targets $\vec{p}_y$ that are estimated in the local oracle situation can offer such information for distillation. Still, even if individual co-occurring contextual cues have been considered in the above-mentioned method, we observe another issue that inhibits the improvement in semantic segmentation. 

However, the impact of the informative soft targets may be overwhelmed by those less informative because they are treated equally in Eq.~\eqref{eq:init_kl_loss}, causing inferior performance as verified in later experiments. Therefore, adjusting the contribution of each element according to the level of information could be beneficial for transferring knowledge in Eq.~\eqref{eq:init_kl_loss}.

In information theory, entropy $\mathcal{H}$ measures the ``amount of information'' in a variable. For the $i$-th element on the pixel-wise prediction $\vec{p}_y \in \mathcal{R}^{[hw, n]}$, $\mathcal{H}^i$ is calculated as:
\begin{equation}
    \label{eq:entropy}
    \mathcal{H}^i = -\sum_{j=1}^{n} \sigma(\vec{p}_y)^{i,j} \cdot \log \sigma(\vec{p}_y)^{i,j} \quad i \in \{1, ..., hw\},
\end{equation}
where $\sigma$ represents the Softmax operation on the second dimension of $\vec{p}_y$. As shown in later experiments, adopting the prediction $\vec{p}_y$ yielded with the oracle contextual prior to estimate $\mathcal{H}$ brings preferable results than $\vec{p}_p$ and $\vec{p}$. 
Then, by incorporating the entropy mask $\mathcal{H} \in \mathcal{R}^{[hw]}$, the distillation loss $\mathcal{L}_{KL}$ introduced in Eq.~\eqref{eq:init_kl_loss} is accordingly updated as:
\begin{equation}
    \label{eq:init_entropy_kl}
    \mathcal{L}_{KL} = \frac{-1}{{\sum_{i=1}^{hw} \mathcal{H}^{i}}} \sum_{i=1}^{hw} \sum_{j=1}^{n} 
                        \mathcal{H}^{i} \sigma(\vec{p}_{y})^{i,j} \log \sigma(\vec{p}_{p})^{i,j}.
\end{equation}

Besides, in semantic segmentation, multiple classes usually exist in a single image, thus the propagated information may still bias towards the classes of the majority. To alleviate this issue, the distillation loss is calculated independently for different categories. Finally, $\mathcal{L}_{KL}$ is formulated as:
\begin{equation}
    \label{eq:final_entropy_kl}
    \mathcal{L}_{KL} = \frac{-1}{n} \sum_{k=1}^{n} \frac {\sum_{i=1}^{hw} \sum_{j=1}^{n} 
                        \mathcal{M}_{k}^{i} \mathcal{H}^{i} \sigma(\vec{p}_{y})^{i,j} \log \sigma(\vec{p}_{p})^{i,j}}
                        {{\sum_{i=1}^{hw} \mathcal{M}_{k}^{i} \mathcal{H}^{i}}}
\end{equation}
where the binary mask $\mathcal{M}_k = (\vec{y} == k)$ indicates the existence of the $k$-th class. 

We note that though it seems to be attainable to directly apply $\mathcal{L}_{KL}$ to regularize the original output $\vec{p}$ instead of $\vec{p}_p$ yielded by the estimated context-aware classifier, experiments in Table~\ref{tab:ablation_alternative_designs} show that applying $\mathcal{L}_{KL}$ to $\vec{p}$ is less effective, certifying the importance of the context-aware classifier. On the other hand, as shown in Table~\ref{tab:ablation_loss}, removing $\mathcal{L}_{KL}$ results in inferior performance, manifesting the fact that both $\mathcal{L}_{KL}$ and context-aware classifier are indispensable.

\mypara{Discussion with self-attention. }
Self-attention (SA) dynamically adapts to different inputs via the weighing matrix obtained by multiplying the key and query vectors yielded by individual inputs. 
Yet, the intrinsic difference is that SA only adjusts features to diverse contexts, leaving the decision boundary in the latent space, \ie, the classifier, untouched, while the proposed method works in another direction by altering the decision boundary according to the contents of various scenarios. As shown in Section~\ref{sec:results}, our method is complimentary to popular SA-based designs, \eg, Swin Transformer~\cite{swin} and OCRNet~\cite{ocr}, by achieving preferable improvements without deprecating the efficiency.

%% file: doc/results.tex
\section{Experiments}

\begin{table*}[!t]
	\centering
				\tabcolsep=0.4cm
				\footnotesize            
				{
					\begin{tabular}{ 
							l
							l
							c  
							c 
							c 
							c 
							c 
							c            }
						\toprule
						& 
						&
						&			
						&\multicolumn{2}{c}{ADE20K} 
						&\multicolumn{2}{c}{Stuff 164K}  
						\\              
						Head
						& Backbone
						& fps
						& \#params. 
						& s.s.
						& m.s.            
						& s.s.
						& m.s.            \\            
						
						\specialrule{0em}{0pt}{1pt}
						\hline
						\specialrule{0em}{1pt}{0pt}
						FCN
						& MobileNet-V2
						& 51.10 
						& 9.82M
						& 19.71
						& 19.56   
						& 15.28
						& 17.01               
						\\     
						
						\rowcolor{gray!25}\ \ + Ours
						& MobileNet-V2
						& 49.46 
						& 10.61M
						& 37.40
						& 39.09  
						& 25.37
						& 27.17                
						\\                 
						
						DeepLab-V3+
						& MobileNet-V2
						& 38.42 
						& 15.35M
						& 34.02
						& 34.82
						& 31.18
						& 32.01                
						\\     
						
						\rowcolor{gray!25}\ \ + Ours
						& MobileNet-V2
						& 36.25
						& 16.13M
						& 39.34
						& 41.28     
						& 34.71
						& 35.81                
						\\                 
						
						\specialrule{0em}{0pt}{1pt}
						\hline
						\specialrule{0em}{1pt}{0pt}
						OCRNet
						& HRNet-W18
						& 14.62 
						& 12.18M
						& 39.32
						& 40.80
						& 31.58
						& 32.34                
						\\                
						
						\rowcolor{gray!25}\ \ + Ours
						& HRNet-W18
						& 14.37 
						& 12.97M
						& 44.47
						& 47.16        
						& 39.12
						& 40.65                 
						\\            
						
						UperNet
						& ResNet-50
						& 24.06 
						& 66.52M
						& 42.05
						& 42.78   
						& 39.86
						& 40.26                 
						\\    
						
						\rowcolor{gray!25}\ \ + Ours
						& ResNet-50
						& 23.37 
						& 67.30M
						& 45.24
						& 46.30
						& 41.26
						& 42.30                 
						\\            
						
						DeepLab-V3+
						& ResNet-50
						& 24.09 
						& 43.69M
						& 43.95
						& 44.93 
						& 40.85
						& 41.49                 
						\\   
						
						\rowcolor{gray!25}\ \ + Ours
						& ResNet-50
						& 23.54
						& 44.48M
						& 46.29
						& 47.56
						& 42.99
						& 43.97           
						\\                
						
						\specialrule{0em}{0pt}{1pt}
						\hline
						\specialrule{0em}{1pt}{0pt}     
						
						OCRNet
						& HRNet-W48
						& 13.78 
						& 70.53M
						& 43.25
						& 44.88
						& 40.40
						& 41.66                 
						\\              
						
						\rowcolor{gray!25}\ \ + Ours
						& HRNet-W48
						& 13.33 
						& 71.32M
						& 45.68
						& 48.13
						& 42.64
						& 43.53                 
						\\           
						
						UperNet
						& ResNet-101
						& 19.65 
						& 85.51M
						& 43.82	
						& 44.85    
						& 41.15
						& 41.51                 
						\\      
						
						\rowcolor{gray!25}\ \ + Ours
						& ResNet-101
						& 19.49
						& 86.30M
						& 46.06
						& 47.74
						& 43.13 
						& 43.84                 
						\\                 
						
						DeepLab-V3+
						& ResNet-101
						& 16.39 
						& 62.68M
						& 45.47	
						& 46.35
						& 42.39
						& 42.96                 
						\\                    
						
						\rowcolor{gray!25}\ \ + Ours
						& ResNet-101
						& 16.03
						& 63.47M
						& 47.25
						& 48.41
						& 44.21
						& 45.10                 
						\\           
						
						\specialrule{0em}{0pt}{1pt}
						\hline
						\specialrule{0em}{1pt}{0pt}                
						
						UperNet
						& Swin-Tiny
						& 20.38
						& 59.94M
						& 44.51 
						& 45.81  
						& 43.83
						& 44.58                 
						\\             
						
						\rowcolor{gray!25}\ \ + Ours
						& Swin-Tiny
						& 19.96
						& 60.73M
						& 46.91
						& 49.03
						& 44.57
						& 45.83                 
						\\             
						
						UperNet
						& Swin-Base$\dagger$
						& 14.63 
						& 121.42M
						& 50.04 
						& 51.66
						& 47.67
						& 48.57                 
						\\              
						
						\rowcolor{gray!25}\ \ + Ours
						& Swin-Base$\dagger$
						& 14.38
						& 122.20M
						& 52.00
						& 53.52
						& 48.26
						& 49.55                 
						\\         
						
						UperNet
						& Swin-Large$\dagger$
						& 10.54
						& 233.96M
						& 52.00
						& 53.50
						& 47.89
						& 48.93                
						\\      
						
						\rowcolor{gray!25}\ \ + Ours
						& Swin-Large$\dagger$
						& 10.44
						& 234.75M
						& 52.87
						& 54.43
						& 48.82 
						& 50.00        
						\\                  
						
						\bottomrule                                   
					\end{tabular}
				}
	\caption{Performance Comparison on ADE20K~\cite{ade20k} and COCO-Stuff 164K~\cite{cocostuff164k}. Single-scale (s.s.) and multi-scale (m.s.) evaluation results are reported, and values of fps (frames per second) are obtained with resolution $512 \times 512$ on a single NVIDIA RTX 2080Ti GPU. Models marked with $\dagger$ are pre-trained on ImageNet-22K following the practice mentioned in~\cite{swin}. }	
	\label{tab:results_ade20k}  		
\end{table*}

\subsection{Implementation}
We adopt two challenging semantic segmentation benchmarks (ADE20K~\cite{ade20k} 
and COCO-Stuff 164K~\cite{cocostuff164k}) in this paper. Models are trained and evaluated on the training and validation sets of these datasets respectively. Results of Cityscapes~\cite{cityscapes} and Pascal-Context~\cite{pascalcontext} are shown in the supplementary due to the page limit. The convolution-based and transformer-based models are investigated by following their default training and testing configurations. Both single- and multi-scale results are reported. Different from the single-scale results that are evaluated on the original size, the multi-scale evaluation conducts inference with the horizontal flipping and scales of [0.5, 0.75, 1.0, 1.25, 1.5, 1.75].  

The projectors $\theta_y$ and $\theta_p$ are both composed of two linear layers ([$2d \times d/2$] $\rightarrow$ [$d/2 \times d$], $d=512$) with an intermediate ReLU layer. The loss weight $\lambda_{KL}$ and the scaling factor $\tau$ for cosine similarity are empirically set to 1 and 15, and they work well in our experiments. Implementations regarding baseline models and benchmarks are based on the default configurations of MMSegmentation~\cite{mmseg2020}, and they are kept intact when implemented with our method.

\subsection{Results}
\label{sec:results}
\mypara{Quantitative results. }
To verify the effectiveness and generalization ability of our proposed method, various decoder heads (FCN~\cite{fcn}, DeepLab-V3+~\cite{deeplabv3+}, UperNet~\cite{upernet}, OCRNet~\cite{ocr}), with different types of backbones, including ResNet~\cite{resnet} and MobileNet~\cite{mobilenetv2}, and Swin Transformer (Swin)~\cite{swin}, are adopted as the baselines. 

The results on ADE20K and COCO-Stuff 164K are shown in Table~\ref{tab:results_ade20k} from which we can observe that the proposed context-aware classifier only introduces about 2\% additional inference time and a few additional parameters to all these baseline models, but decent performance gain has been achieved on both two challenging benchmarks, including the model implemented with powerful transformer Swin-Large, reaching impressive performance without compromising the efficiency. 
It is worth noting that, the improvement is not originated from the newly introduced parameters, because our method even helps smaller models beat the larger ones with much more parameters, such as DeepLabV3+ (Res-50)  \textit{v.s.} OCRNet (HR-48) and UperNet (Swin-Base$\dagger$) \textit{v.s.} UperNet (Swin-Large$\dagger$). 

\begin{figure}[!t]
	\centering
	\begin{minipage}    {0.19\linewidth}
		\centering
		\includegraphics [width=1\linewidth,height=0.7\linewidth] 
		{figure/img_gt/ADE_val_00000178.jpg}
	\end{minipage}    
	\begin{minipage}  {0.19\linewidth}
		\centering
		\includegraphics [width=1\linewidth,height=0.7\linewidth] 
		{figure/img_gt/ADE_val_00000222.jpg}
	\end{minipage}    
	\begin{minipage}  {0.19\linewidth}
		\centering
		\includegraphics [width=1\linewidth,height=0.7\linewidth] 
		{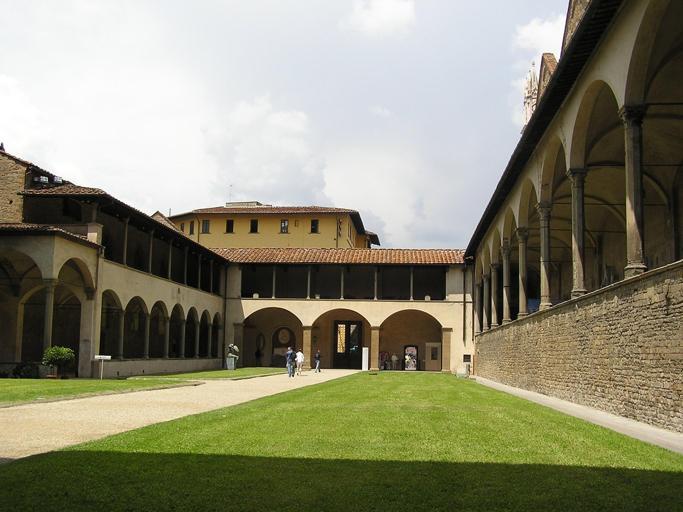}
	\end{minipage}        
	\begin{minipage}  {0.19\linewidth}
		\centering
		\includegraphics [width=1\linewidth,height=0.7\linewidth] 
		{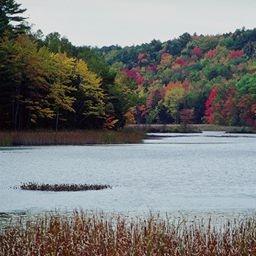}
	\end{minipage}    
	\begin{minipage}  {0.19\linewidth}
		\centering
		\includegraphics [width=1\linewidth,height=0.7\linewidth]  
		{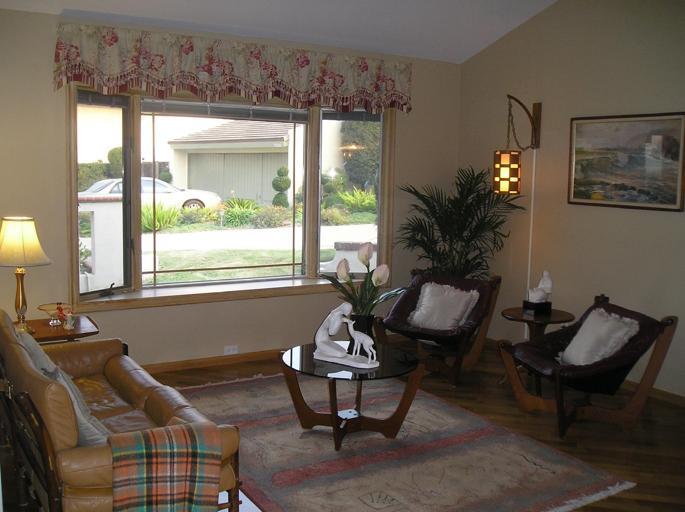}
	\end{minipage}    
	
	\begin{minipage}  {0.19\linewidth}
		\centering
		\includegraphics [width=1\linewidth,height=0.7\linewidth] 
		{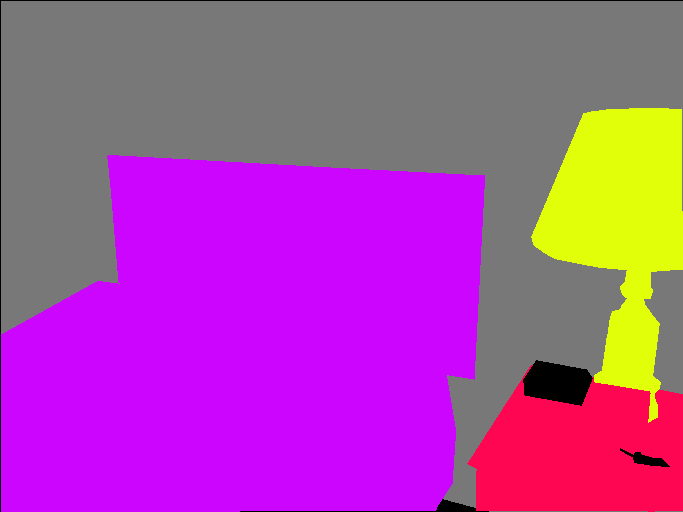}
	\end{minipage}    
	\begin{minipage}  {0.19\linewidth}
		\centering
		\includegraphics [width=1\linewidth,height=0.7\linewidth] 
		{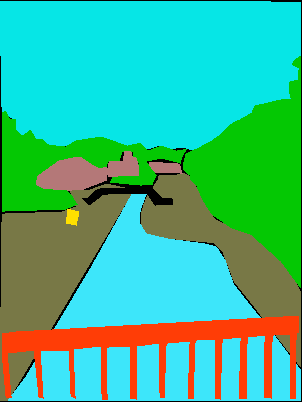}
	\end{minipage}      
	\begin{minipage}  {0.19\linewidth}
		\centering
		\includegraphics [width=1\linewidth,height=0.7\linewidth] 
		{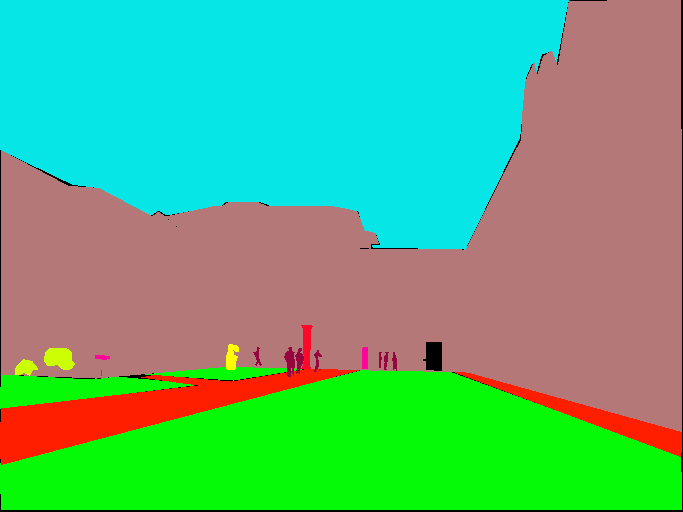}
	\end{minipage}    
	\begin{minipage}  {0.19\linewidth}
		\centering
		\includegraphics [width=1\linewidth,height=0.7\linewidth] 
		{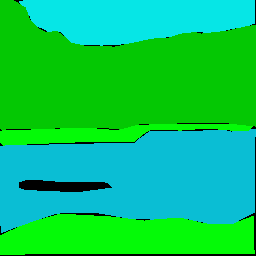}
	\end{minipage}    
	\begin{minipage}  {0.19\linewidth}
		\centering
		\includegraphics [width=1\linewidth,height=0.7\linewidth] 
		{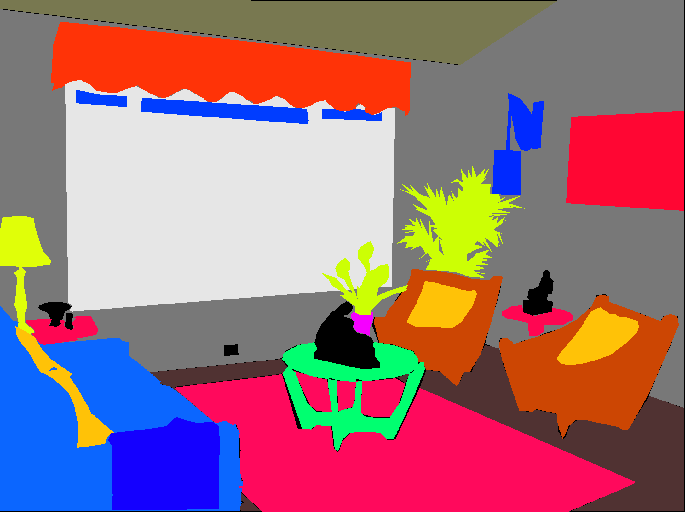}
	\end{minipage}        
	
	\begin{minipage}  {0.19\linewidth}
		\centering
		\includegraphics [width=1\linewidth,height=0.7\linewidth] 
		{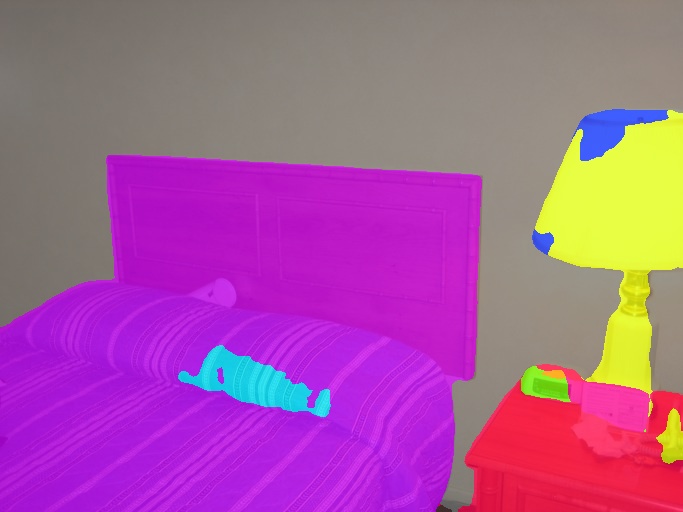}
	\end{minipage}    
	\begin{minipage}  {0.19\linewidth}
		\centering
		\includegraphics [width=1\linewidth,height=0.7\linewidth] 
		{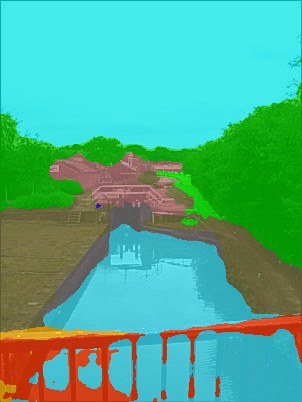}
	\end{minipage}       
	\begin{minipage}  {0.19\linewidth}
		\centering
		\includegraphics [width=1\linewidth,height=0.7\linewidth] 
		{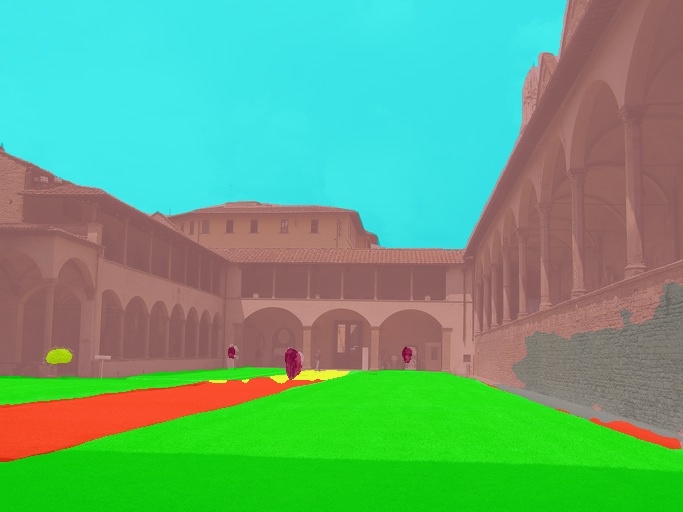}
	\end{minipage}    
	\begin{minipage}  {0.19\linewidth}
		\centering
		\includegraphics [width=1\linewidth,height=0.7\linewidth] 
		{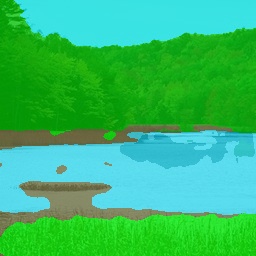}
	\end{minipage}    
	\begin{minipage}  {0.19\linewidth}
		\centering
		\includegraphics [width=1\linewidth,height=0.7\linewidth] 
		{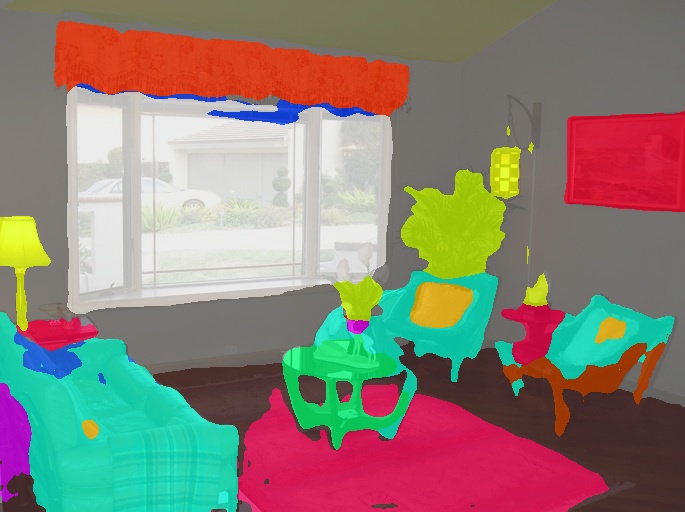}
	\end{minipage}      
	
	\begin{minipage}  {0.19\linewidth}
		\centering
		\includegraphics [width=1\linewidth,height=0.7\linewidth] 
		{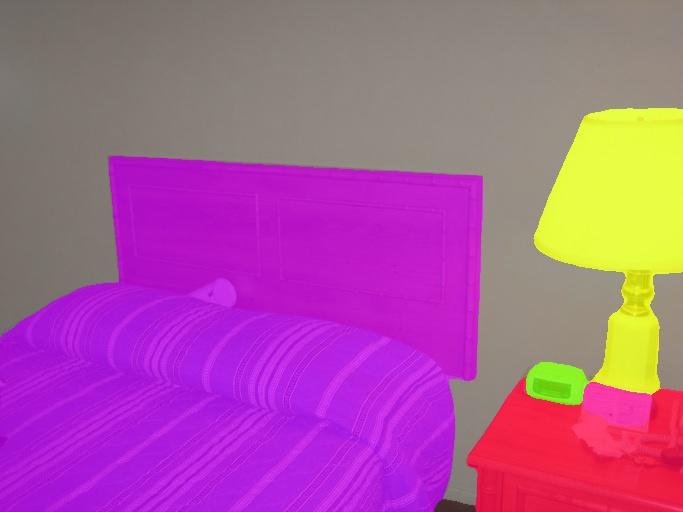}
	\end{minipage}    
	\begin{minipage}  {0.19\linewidth}
		\centering
		\includegraphics [width=1\linewidth,height=0.7\linewidth] 
		{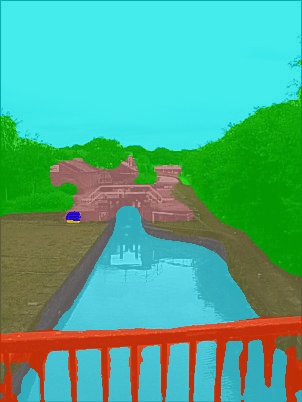}
	\end{minipage}    
	\begin{minipage}  {0.19\linewidth}
		\centering
		\includegraphics [width=1\linewidth,height=0.7\linewidth] 
		{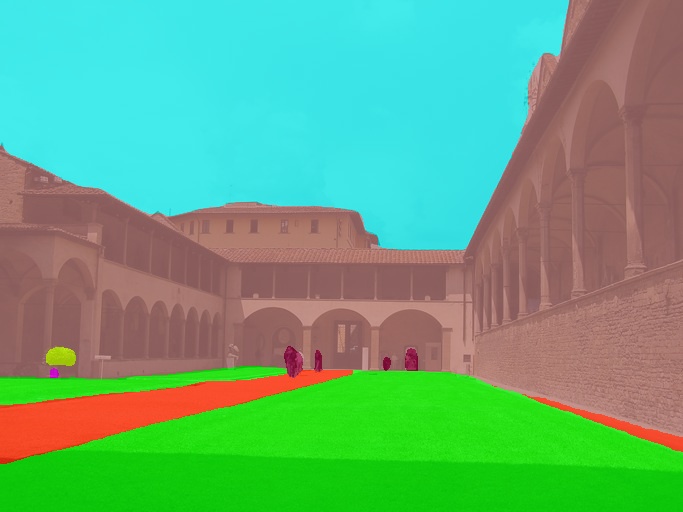}
	\end{minipage}    
	\begin{minipage}  {0.19\linewidth}
		\centering
		\includegraphics [width=1\linewidth,height=0.7\linewidth] 
		{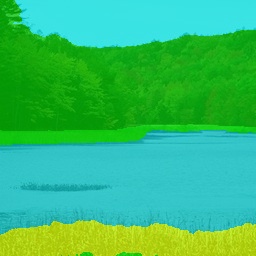}
	\end{minipage}    
	\begin{minipage}  {0.19\linewidth}
		\centering
		\includegraphics [width=1\linewidth,height=0.7\linewidth]  
		{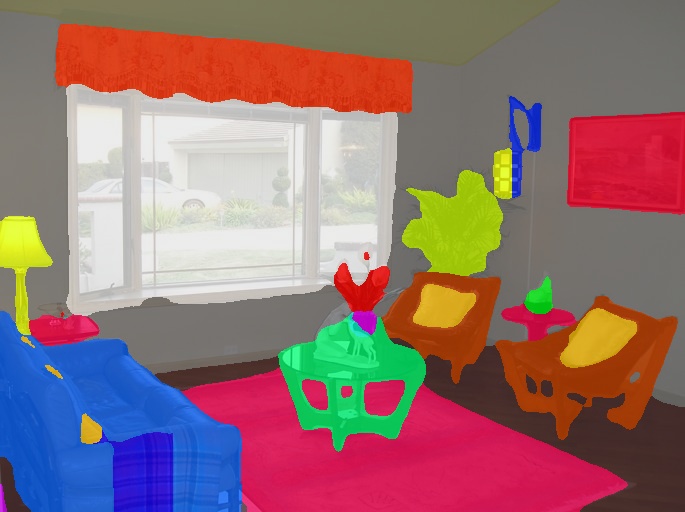}
	\end{minipage}    
	
	\caption{Visual illustrations from top to bottom are from input images, ground-truth, baseline and baseline+ours. Black regions are ignored during testing. }
	\label{fig:vis_example}
\end{figure}

\begin{figure}[!t]
	\centering
	\begin{minipage}   {0.24\linewidth}
		\centering
		\includegraphics [width=1\linewidth] 
		{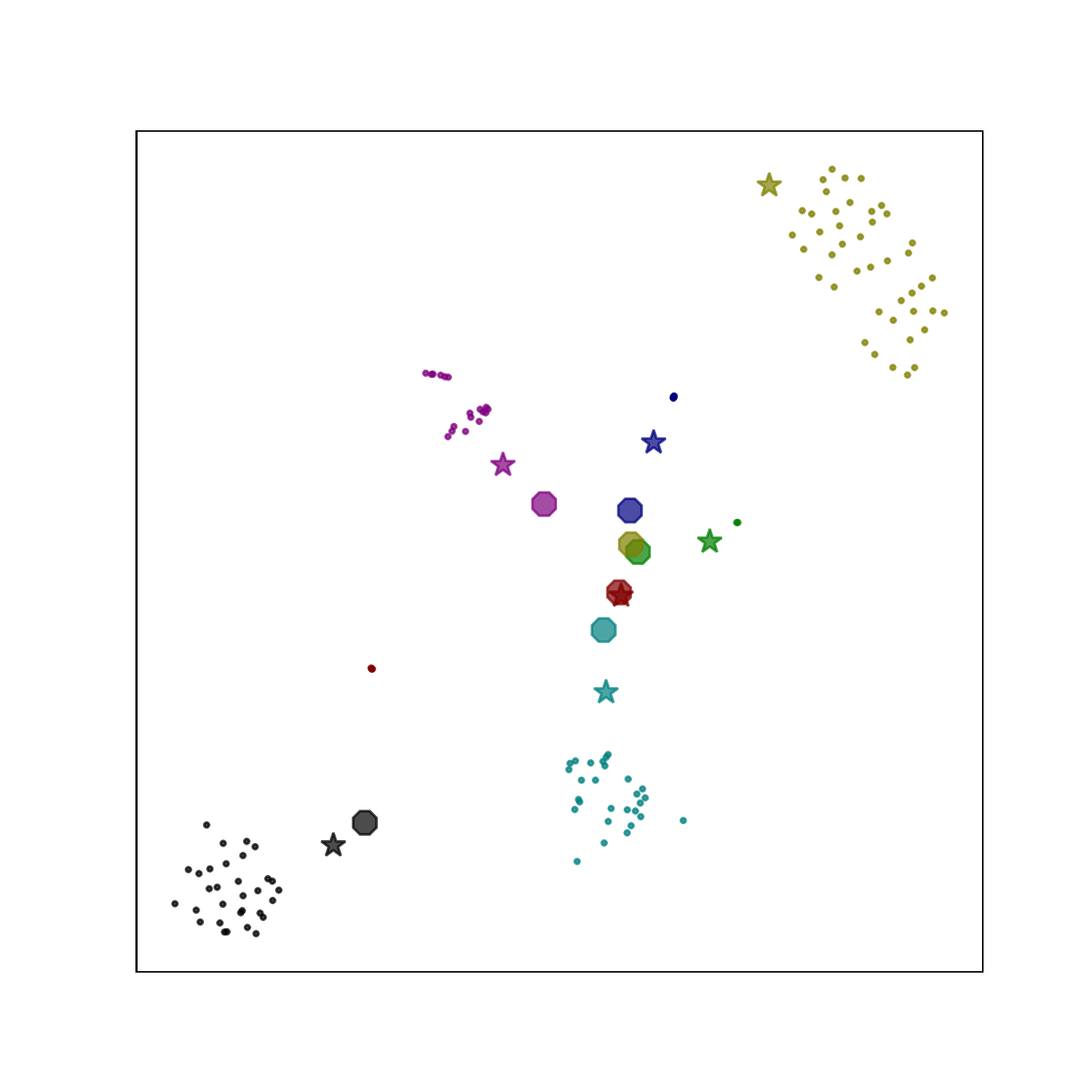}
	\end{minipage}    
	\begin{minipage}   {0.24\linewidth}
		\centering
		\includegraphics [width=1\linewidth] 
		{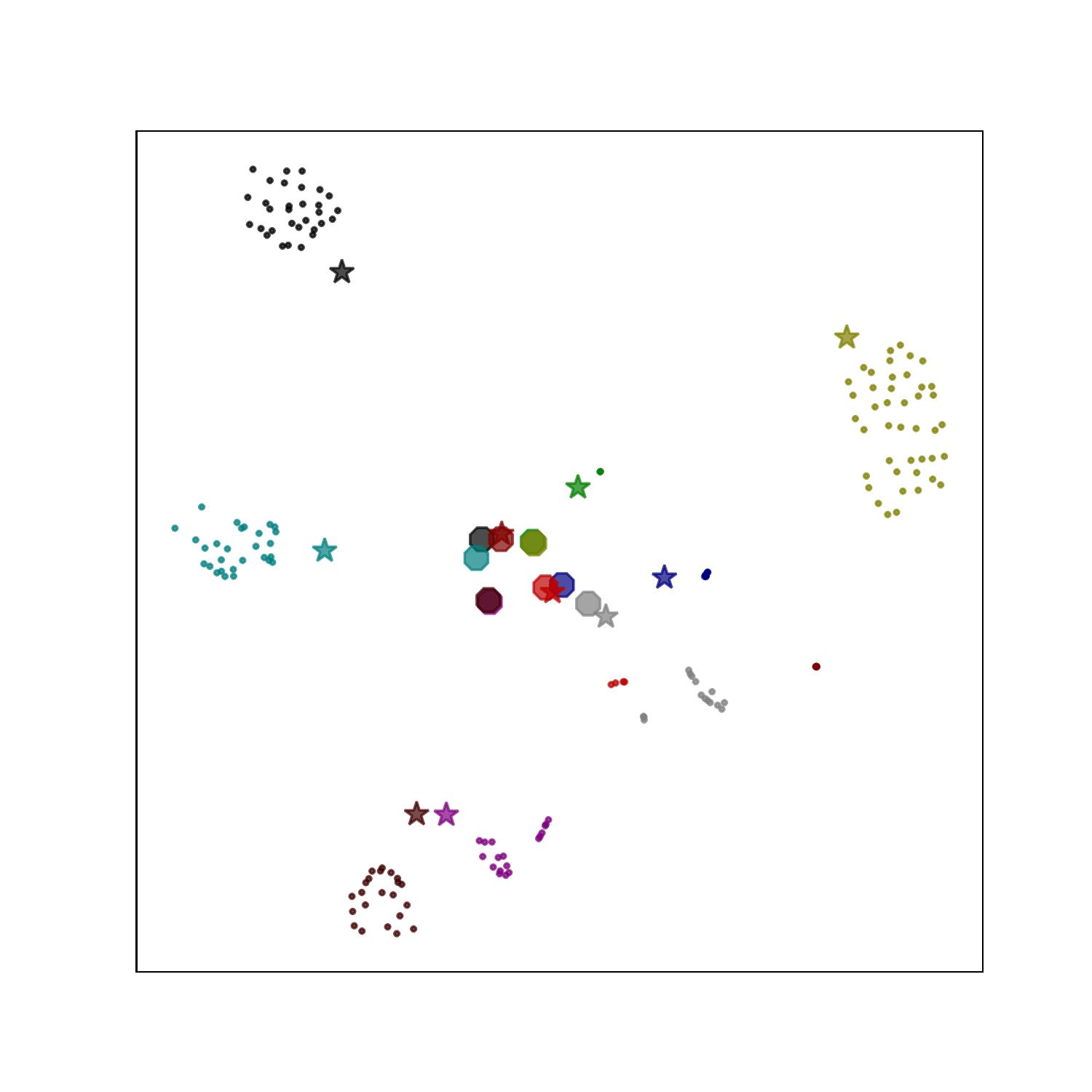}
	\end{minipage}        
	\begin{minipage}   {0.24\linewidth}
		\centering
		\includegraphics [width=1\linewidth] 
		{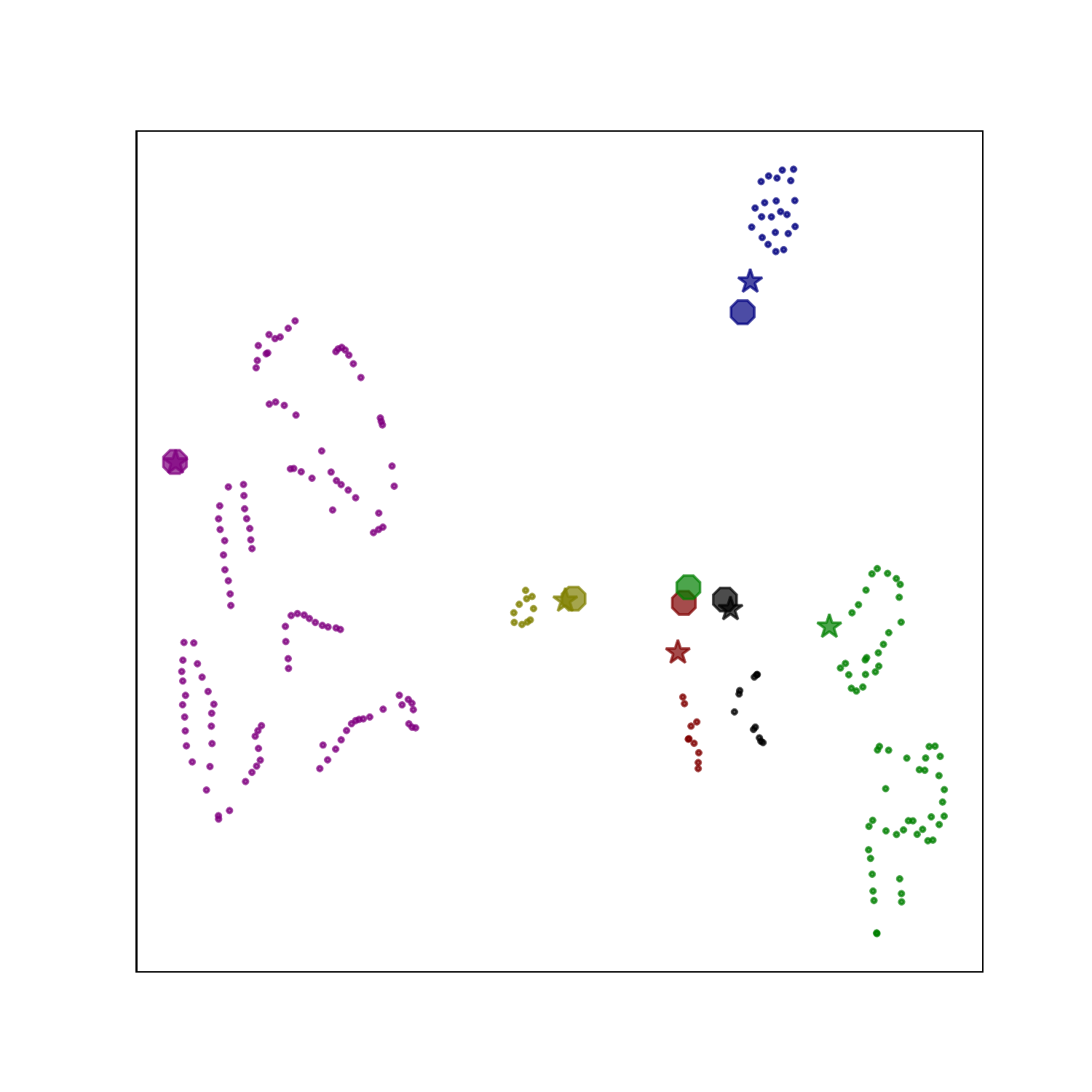}
	\end{minipage}     
	\begin{minipage}   {0.24\linewidth}
		\centering
		\includegraphics [width=1\linewidth] 
		{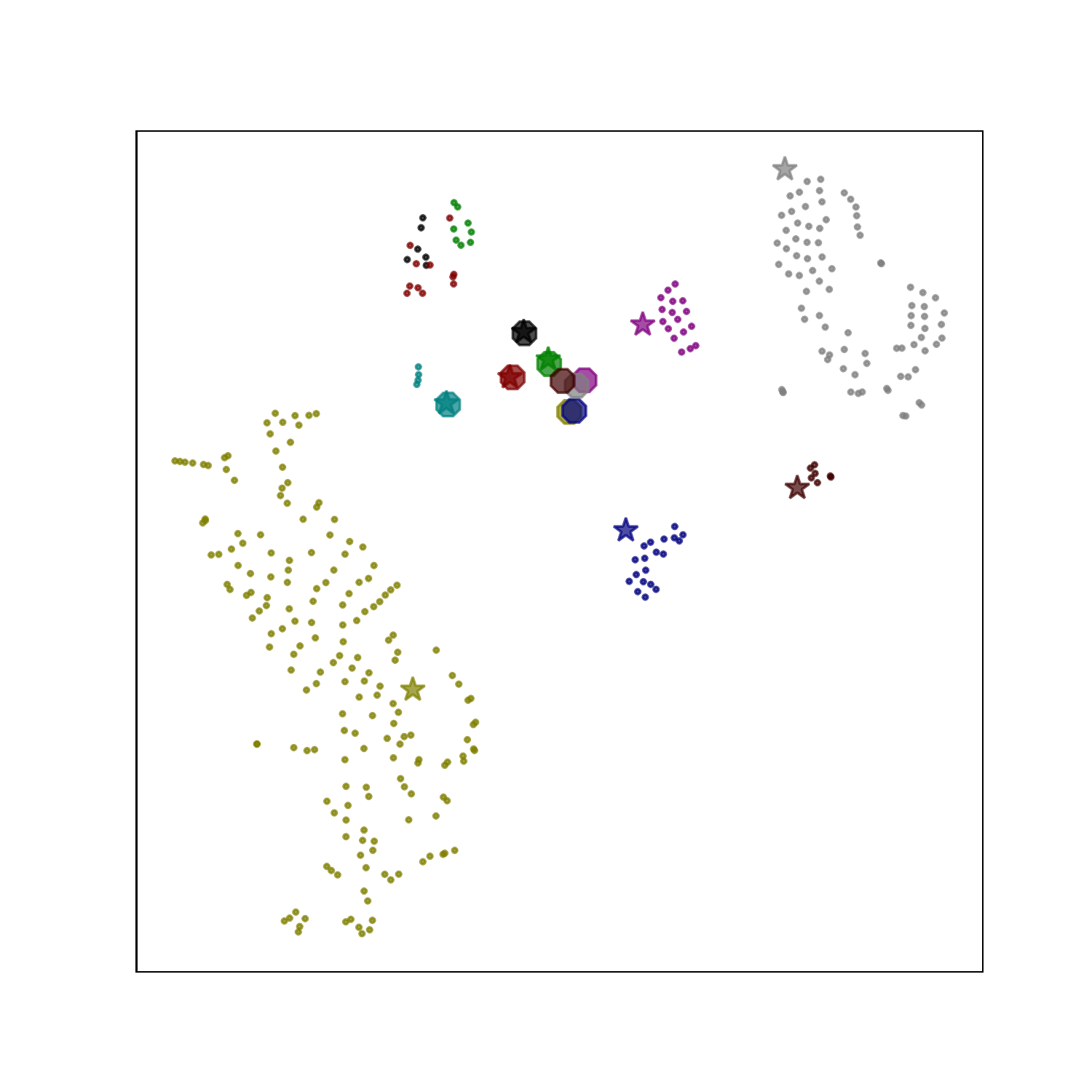}
	\end{minipage}           
	\caption{Results of t-SNE. Categories are represented in different colors. Small dots are feature vectors, large circles are the weights of the original classifier, and stars are the weights of the approximated context-aware classifier. }
	\label{fig:post_tsne}
\end{figure}

\mypara{Qualitative results. }
Predicted masks are shown in Figure~\ref{fig:vis_example} where the ones yielded with our proposed method are more visually attractive. 
Besides, for facilitating the understanding, t-SNE results are demonstrated in Figure~\ref{fig:post_tsne}. It can be observed that, with the proposed learning scheme, the estimated context-aware classifiers are more semantically representative to different individuals by effectively rectifying the original classifier with necessary contextual information.

\subsection{Ablation Study}
\label{sec:ablation_study}
In this section, experimental results are presented to investigate the effectiveness of each component of our proposed method. The ablation study is conducted on ADE20K, and the baseline model is UperNet with Swin-Tiny.

\mypara{Effects of different loss combinations. }
$\mathcal{L}^{ce}$ supervises the original classifier's prediction $\vec{p}$ that is used for generating the estimated context-aware prototypes $\mathcal{C}_{p}$. Since $\mathcal{A}_{p}$ is an approximation of the oracle one $\mathcal{A}_{y}$ yielded with the ground-truth label $\vec{y}$, the supervisions on $\vec{p}$ and   $\mathcal{A}_{y}$ are both essential. To examine the effects of individual losses, experimental results are in Table~\ref{tab:ablation_loss}.
It can be observed from (b) that, without $\mathcal{L}^{ce}$, $\mathcal{L}^{ce}_{y}$ and $\mathcal{L}_{KL}$, merely supervising $\vec{p}_{p}$ even worsens the baseline's performance (a), and the comparison between (b) and (c) tells the importance of $\mathcal{L}^{ce}$ that supervises $\vec{p}$. The other experiments show the necessities of $\mathcal{L}_{KL}$ and $\mathcal{L}^{ce}_{y}$. Specifically, since $\mathcal{L}_{KL}$ encourages $\mathcal{A}_{p}$ to mimic $\mathcal{A}_{y}$, though the comparison between (c) \& (d) implies additionally optimizing the prediction of the oracle case is beneficial, (d) is still inferior to (f) that incorporates $\mathcal{L}_{KL}$. On the other hand, without $\mathcal{L}^{ce}_{y}$ that ensures the validity of the prediction in the oracle case, the result of (e) is clearly lower than that of (f). Moreover, the sensitivity analysis on the loss weight $\lambda_{KL}$ for $\mathcal{L}_{KL}$ is demonstrated by the results of (f)-(h), and setting $\lambda_{KL}$ to 1 is found satisfactory.

\begin{table}[!t]
	\centering
	\tabcolsep=0.4cm
	\footnotesize            
	{
            \begin{tabular}{ 
					l
					c      }
				\toprule
				Loss Function
				& mIoU
				\\            
				
				\specialrule{0em}{0pt}{1pt}
				\hline
				\specialrule{0em}{1pt}{0pt}
				(a) $\mathcal{L}=\mathcal{L}^{ce}$ (Baseline)
				& 44.51
				\\            
				
				(b) $\mathcal{L}=\mathcal{L}^{ce}_{p}$
				& 44.06
				\\               
				
				(c) $\mathcal{L}=\mathcal{L}^{ce} + \mathcal{L}^{ce}_{p}$
				& 45.14
				\\                   
				
				(d) $\mathcal{L}=\mathcal{L}^{ce} + \mathcal{L}^{ce}_{p} + \mathcal{L}^{ce}_{y}$
				& 45.74
				\\                    
				
				(e) $\mathcal{L}=\mathcal{L}^{ce} + \mathcal{L}^{ce}_{p} + \mathcal{L}_{KL}$
				& 45.23
				\\              
				
				\rowcolor{gray!25}
				(f) $\mathcal{L}=\mathcal{L}^{ce} + \mathcal{L}^{ce}_{p} + \mathcal{L}^{ce}_{y} + \mathcal{L}_{KL}$
				& 46.91
				\\                              
				
				(g) $\mathcal{L}=\mathcal{L}^{ce} + \mathcal{L}^{ce}_{p} + \mathcal{L}^{ce}_{y} + 0.1 \cdot \mathcal{L}_{KL}$
				& 45.88
				\\

				(h) $\mathcal{L}=\mathcal{L}^{ce} + \mathcal{L}^{ce}_{p} + \mathcal{L}^{ce}_{y} + 10 \cdot \mathcal{L}_{KL}$
				& 46.08
				\\                
				
				\bottomrule                                   
			\end{tabular}
	}
    \caption{Ablation study on different loss combinations. }
	\label{tab:ablation_loss}     
\end{table}     

\begin{table}[!t]
	\centering
	\tabcolsep=0.4cm
	\footnotesize            
	{
            \begin{tabular}{ 
					l
					c      }
				\toprule
				Loss Function
				& mIoU
				\\            
				
				\specialrule{0em}{0pt}{1pt}
				\hline
				\specialrule{0em}{1pt}{0pt}
				(a) w/o KL
				& 45.74
				\\            
				
				(b) Vanilla KL
				& 45.72
				\\            
				
				(c) Entropy KL
				& 45.99
				\\           
				
				(d) Class-wise KL
				& 46.10
				\\                
				
				\rowcolor{gray!25}
				(e) Class-wise Entropy KL
				& 46.91
				\\          
				
				\specialrule{0em}{0pt}{1pt}
				\hline
				\specialrule{0em}{1pt}{0pt}            
				
				(1) Class-wise Entropy KL (Est.)
				& 46.58
				\\       
				
				(2) Class-wise Entropy KL (Ori.)
				& 46.22
				\\                
				\bottomrule                                   
			\end{tabular}
	}
	\caption{Ablation study on the designs for KL loss.  }  
	\label{tab:ablation_klloss}    
\end{table}     

\mypara{Different forms of KL loss. }
Section~\ref{sec:learning_cwc} introduces the vanilla KL loss that encourages the model to learn to form the context-aware classifier. 
To alleviate the information bias and further exploit hidden useful cues, we propose an alternative form that leverages the class-wise entropy. To show the effectiveness of the proposed design on $\mathcal{L}_{KL}$, results are shown in Table~\ref{tab:ablation_klloss} where Exp. (a) is the same as Exp. (d) in Table~\ref{tab:ablation_loss} without KL loss.
Besides, different from Exps.~(d)-(e) whose entropy mask is obtained from $\vec{p}_y$,
entropy masks in Exps.~(1)-(2) are estimated by $\vec{p}_p$ and $\vec{p}$ respectively.

In Table~\ref{tab:ablation_klloss}, the vanilla KL loss achieves comparable to Exp. (a) without KL loss, and the entropy-based KL in Exp. (c) also incrementally improves Exp. (a) because the information of the majority may still overwhelm the others. Instead, by applying class-wise calculation to (b), improvement is obtained in Exp. (d), since it helps alleviate the imbalance between different classes. Furthermore, to tackle the information bias, informative cues are better exploited in Exp. (e) by incorporating the entropy estimation with the class-wise KL, achieving persuasive performance. Last, Exp. (1) and Exp. (2) prove that the oracle predictions $\vec{p}_y$ are more favorable than $\vec{p}_p$ (Est.) and $\vec{p}$ (Ori.) for estimating the entropy mask used in Eq.~\eqref{eq:entropy}.

\begin{table}[!t]
	\centering
	\tabcolsep=0.4cm
	\footnotesize            
	{
		\begin{tabular}{ 
				l
				c      }
			\toprule
			Classifier
			& mIoU
			\\            
			
			\specialrule{0em}{0pt}{1pt}
			\hline
			\specialrule{0em}{1pt}{0pt}
			(a) Original (Dot) 
			& 44.51
			\\                 
			
			(b) Original (Cos) 
			& 43.89
			\\                    
			
			(c) Original (Dot) + Context (Dot)
			& 45.42
			\\                  
			
			\rowcolor{gray!25}
			(d) Original (Dot) + Context (Cos)
			& 46.91
			\\            
			
			(e) Original (Cos) + Context (Cos)
			& 46.39
			\\       
			
			\specialrule{0em}{0pt}{1pt}
			\hline
			\specialrule{0em}{1pt}{0pt}            
			
			(1) Exp. (d) with ($\tau = 5$)
			& 45.39
			\\                 
			
			(2) Exp. (d) with  ($\tau = 10$)
			& 46.78
			\\                 
			
			(3) Exp. (d) with ($\tau = 20$)
			& 46.26
			\\           
			
			\bottomrule                                   
		\end{tabular}
	}
	\caption{Ablation study on the cosine similarity and dot product on the original and the proposed context-aware classifiers. Exps. (b), (d) and (e) are with $\tau=15$. }  
	\label{tab:ablation_cosine} 
\end{table}

\mypara{The necessity of cosine similarity. }
In segmentation models, the original classifier $\mathcal{C} \in \mathcal{R}^{[n, d]}$ applies dot product on the features $\vec{f} \in \mathcal{R}^{[hw, d]}$ yielded by the feature generator to get the output $\vec{p} = \vec{f} \times \mathcal{C}^\top = |\vec{f}||\mathcal{C}| \mathop{\cos}(\vec{f}, \mathcal{C})$. 
However, the proposed context-aware classifier yields predictions via cosine-similarity 
$\vec{p}_a 
= \tau \cdot \mathop{\cos}(\vec{f}, \mathcal{A}_*) 
= \tau \cdot \eta(\vec{f}) \times \eta(\mathcal{A}_*)^\top$ ($* \in \{y, p\}$).
The difference is that, cosine similarity focuses on the angle between two vectors, while dot product considers both angle and magnitudes.

Though both cosine similarity and dot product seem plausible, since the norms $|\vec{f}|$ and $|\mathcal{C}|$ are not bounded, extreme values may occur and hinder the optimization process of the context-aware classifier to proceed normally as that with the original classifier. On the contrary, the instability issues caused by the magnitudes of dynamically imprinted classifier weights $\mathcal{A}_y$ and $\mathcal{A}_p$ can be alleviated by applying L-2 normalization in the cosine function.

\begin{table}[!t]
	\centering
	\tabcolsep=0.5cm
	\footnotesize            
	{
        \begin{tabular}{ 
              l
              c           }
             \toprule
             Model
             & mIoU
             \\            

             \specialrule{0em}{0pt}{1pt}
             \hline
             \specialrule{0em}{1pt}{0pt}
             Baseline
             & 44.51
             \\

             (a) $\mathcal{A}_* = \mathcal{C}_*$
             & 44.08
             \\               

             (b) $\mathcal{A}_* = \mathcal{C}_* + \mathcal{C}$
             & 44.13
             \\

             (c) $\mathcal{A}_* = \theta_*(\mathcal{C}_*)$
             & 46.21
             \\   

             (d) $\mathcal{A}_* = \theta_*(\mathcal{C}_* + \mathcal{C})$
             & 46.53
             \\    

             \rowcolor{gray!25}
             (\#)  $\mathcal{A}_* = \theta_*(\mathcal{C}_* \oplus \mathcal{C})$
             & 46.91
             \\                 

             (e)  $\mathcal{A}_* = \theta_*(\mathcal{C}_* \oplus \mathcal{C}) + \mathcal{C}$
             & 45.94
             \\              

             (f) $\mathcal{A}_y =  \mathcal{C}_y, \mathcal{A}_p = \theta_p(\mathcal{C}_p \oplus \mathcal{C})$

             & 46.01
             \\

             (g) $\mathcal{A}_y = \theta_y(\mathcal{C}_y \oplus \mathcal{C}), \mathcal{A}_p = \mathcal{C}_p$
             & 44.55
             \\                           

             (h) $\mathcal{A}_y =  \mathcal{C}, \mathcal{A}_p = \theta_p(\mathcal{C}_p \oplus \mathcal{C})$
             & 45.69 
             \\                

             (i) $\mathcal{A}_y = \theta_y(\mathcal{C}_y \oplus \mathcal{C}), \mathcal{A}_p = \mathcal{C}$
             & 44.62
             \\                 
             \bottomrule                                   
         \end{tabular}
	}
    \caption{
	Ablation study on alternative designs for yielding the context-aware classifier.
	All models except for the baseline are optimized with $\mathcal{L}_{KL}$. 
}
    \label{tab:ablation_alternative_designs}   
\end{table}

Experimental results are shown in Table~\ref{tab:ablation_cosine} where the context-aware classifier implemented with cosine similarity achieves favorable results while the dot product is better for the original classifier.
This discrepancy may be related to their formation processes.
The original one is shared by all samples, but the weights of the context-aware classifier are dynamically imprinted, thus the former may find an optimal magnitude that generalizes well to a universal distribution throughout the training process. Since magnitudes provide additional information regarding different categorical distributions, dot-product works better on the original classifier. 

Differently, because the approximated context-aware classifier is generated individually, the overall categorical magnitudes may be dominated by the features with large magnitudes, overwhelming those with smaller magnitudes. Furthermore, even for the same class, the feature magnitude will change because of the varying co-occurring stuff and things in different images. Therefore, cosine similarity simply ignores the unstable magnitudes but instead focuses on the inter-class relations, bringing better results to the context-aware classifier. 
In addition, the sensitivity analysis regarding different values of scaling factor $\tau$ shows that the results of $\tau=\{5, 10, 20\}$ are inferior to Exp. (d) with $\tau=15$. Therefore, We set $\tau$ to 15 in all experiments.

\mypara{Alternative designs for yielding context-aware classifier. }
As shown in Eqs.~\eqref{eq:oracle_cls} and \eqref{eq:estimate_cls} in Section~\ref{sec:method}, the oracle and the estimated context-aware classifiers $\mathcal{A}_*$ ($*$ is the placeholder for $y$ and $p$) are generated by applying the projectors $\theta_*$ to the concatenation of the estimated contextual prototypes $\mathcal{C}_*$ and the weights $\mathcal{C}$ of the original classifier, \ie, $\mathcal{A}_* = \theta_*(\mathcal{C}_* \oplus \mathcal{C})$. There are several other design options and results are shown in Table~\ref{tab:ablation_alternative_designs}.

Concretely, (a) means both $\mathcal{A}_y$ and $\mathcal{A}_p$ are simply formed by the oracle and estimated semantic prototypes, and (b) adds the weights of the original classifier as a residue. However, both (a) and (b) cause performance deduction because the estimated prototypes $\mathcal{C}_p$ may deliver irrelevant or even erroneous messages without any processing. Differently, adding a projector to the estimated prototypes $\mathcal{C}_p$ is helpful, as verified by (c), since the projector keeps the essence and screens the noise from $\mathcal{C}_p$.
 Moreover, introducing the information contained in the original classifier is found conducive, as shown by (d) and (\#), and (\#) shows that the concatenation operation is more effective than simply adding the prototypes. But, adding the original ones as a residue in (e) degrades the performance because it may lead to a trivial solution that is to simply skip the projector, which is easier for optimization. The last four results of (f)-(i) are inferior to that of (\#), manifesting the necessity of adopting (\#) to yield both $\mathcal{A}_y$ and $\mathcal{A}_p$. Also, the comparison between (\#) and (i) shows that removing the estimated context-classifier may cause significant performance deduction. The discussion of the projector's structure is in the supplementary file.

\mypara{Impact on model efficiency. }
Our proposed method is effective yet efficient since, during inference, it only introduces an additional lightweight projector and several simple matrix operations to the original model. To comprehensively study the impacts brought to the model efficiency, frames per second (fps) and GPU memory consumption (Mem) obtained in higher input resolutions, \ie, $1280\times1280$ and $2560 \times 2560$, are presented in Table~\ref{tab:ablation_efficiency} from which we can observe that only minor negative impacts are brought to the baseline models, even with the high input resolutions. 

\begin{table}[!t]
    \centering
    \tabcolsep=0.12cm
    \footnotesize            
    {
        \begin{tabular}{ 
           l
           c    
           c
           c    
           c
           c
           c               }
           
            \toprule
			&\multicolumn{2}{c}{$512 \times 512$} 			
			&\multicolumn{2}{c}{$1280 \times 1280$}  
			&\multicolumn{2}{c}{$2560 \times 2560$}  
			
            \\
            Model
            & fps 
            & Mem                
            & fps 
            & Mem                
            & fps 
            & Mem                
		    \\            

            \specialrule{0em}{0pt}{1pt}
            \hline
            \specialrule{0em}{1pt}{0pt}
            Baseline
            & 20.38
            & 2794 
            & 4.13
            & 5650 
            & 1.04
            & 10286          
            \\                 
         
            \rowcolor{gray!25}
            Ours
            & 19.96
            & 2796  
            & 4.05
            & 5654
            & 1.02
            & 10290            
            \\    
            
            \rowcolor{gray!45}
            $\triangle$
            & -2.05\%
            & +0.07\%
            & -1.94\%
            & +0.07\%  
            & -2.00\%
            & +0.04\% 
            \\                
            
            \bottomrule                                   
        \end{tabular}
    }
    \caption{Comparison regarding FLOPs and frames per second (fps) and GPU memory usage (Mem) in different input resolutions. $\triangle$ denotes the relative change. The baseline model is UperNet+Swin-Tiny, and the results of fps are obtained on a single NVIDIA RTX 2080Ti GPU. }
    \label{tab:ablation_efficiency} 
\end{table}

%% file: doc/conclusion.tex
\section{Concluding Remarks}
In this paper, we present learning context-aware classifier as a means to capture and leverage useful contextual information in different samples, improving the performance by dynamically forming specific descriptors for individual latent distributions. 
The feasibility is verified by an oracle case and the model is then required to approximate the oracle during training, so as to adapt to diverse contexts during testing. 
Besides, an entropy-aware distillation loss is proposed to better mine those under-exploited informative hints. In general, our method can be easily applied to generic segmentation models, boosting both small and large ones with favorable improvements without compromising the efficiency, manifesting the potential for being a general yet effective module for semantic segmentation.